\documentclass{Interspeech2024}

\usepackage{amsmath}
\usepackage{algorithmic}
\usepackage{algorithm}
\usepackage{subfig}

\interspeechcameraready

\title{Finding Task-specific Subnetworks in Multi-task Spoken Language Understanding Model}

\name[affiliation={1}]{Hayato}{Futami}
\name[affiliation={2}]{Siddhant}{Arora}
\name[affiliation={1}]{Yosuke}{Kashiwagi}
\name[affiliation={1}]{Emiru}{Tsunoo}
\name[affiliation={2}]{Shinji}{Watanabe}

\address{
  $^1$Sony Group Corporation, Japan \\
  $^2$Carnegie Mellon University, USA}
\email{Hayato.Futami@sony.com}


\begin{document}

\maketitle

\begin{abstract}
Recently, multi-task spoken language understanding (SLU) models have emerged, designed to address various speech processing tasks.
However, these models often rely on a large number of parameters.
Also, they often encounter difficulties in adapting to new data for a specific task without experiencing catastrophic forgetting of previously trained tasks.
In this study, we propose finding task-specific subnetworks within a multi-task SLU model via neural network pruning.
In addition to model compression, we expect that the forgetting of previously trained tasks can be mitigated by updating only a task-specific subnetwork.
We conduct experiments on top of the state-of-the-art multi-task SLU model ``UniverSLU'', trained for several tasks such as emotion recognition (ER), intent classification (IC), and automatic speech recognition (ASR).
We show that pruned models were successful in adapting to additional ASR or IC data with minimal performance degradation on previously trained tasks.
\end{abstract}

\section{Introduction}
Many recent studies on language and speech processing have been advancing toward unified models that can solve a wide range of tasks.
Large Language Models (LLMs) \cite{Radford19-GPT2} are epitomes of those.
LLMs have the ability to perform various natural language processing (NLP) tasks, instructed by prompts that describe the task.
In the field of speech processing, SpeechPrompt \cite{Chang22-SP, Chang23-SP2} has proposed prompt tuning on Generative Spoken Language Model (GSLM) \cite{Lakhotia21-GSLM} to perform a wide variety of spoken language understanding (SLU) tasks.
SLU, which we focus on in this study, covers understanding semantics, paralinguistics, and content from speech \footnote{We use the term SLU in a broader sense, following \cite{Chang22-SP, Chang23-SP2, Arora23-USLU}}.
We include a few representative SLU tasks in this study, such as intent classification (IC), emotion recognition (ER), and automatic speech recognition (ASR).
Recently, UniverSLU has been proposed as a universal spoken language understanding (SLU) model, which fine-tunes Whisper \cite{Radford22-Whisper} by extending Whisper-style task specifier to various SLU tasks \cite{Arora23-USLU}.
UniverSLU has shown superior performances compared to the state-of-the-art task-specific models and prior unified models.
More recently, several studies have been working on extending LLMs to process audio input, performing some SLU tasks \cite{Wang23-LauraGPT, Tang23-SALMONN, Gong23-LTUAS}.

Although such unified multi-task models are promising in performance and usability, they often suffer from large model sizes.
Another problem is that when new data for a specific task become available, additional training on the task can lead to catastrophic forgetting \cite{Goodfellow14-CF} of other tasks trained before.
This is addressed by continual learning, which has been getting attention in ASR \cite{Chang21-TLL, Yang22-OCL, Takashima22-UOE, Diwan22-DC, Eeckt23-UA} and SLU \cite{Cappellazzo23-SKDSLU, MYang24-EICLSLU}.
Continual learning is especially in demand for large-scale speech foundation models, as they require high retraining cost \cite{Wu24-CLLLM}.

To solve the above two issues, we propose a network pruning method to find task-specific subnetworks in a multi-task SLU model.
Pruning has been applied to ASR and SLU to reduce memory footprint and speedup in inference \cite{Lai21-PARP, Ding22-AL, Peng23-SP, Yang22-OSD}.
A recent study on multi-lingual ASR \cite{Yang23-LAP} has succeeded in finding language-specific subnetworks via Lottery Ticket Hypothesis \cite{Frankle18-LTH} -based pruning.
In this study, we extend it to multi-task SLU, where pruning is applied to identify task-specific subnetworks defined as pathways on the model.
\begin{figure}[t]
\centering
\includegraphics[width=0.95\columnwidth]{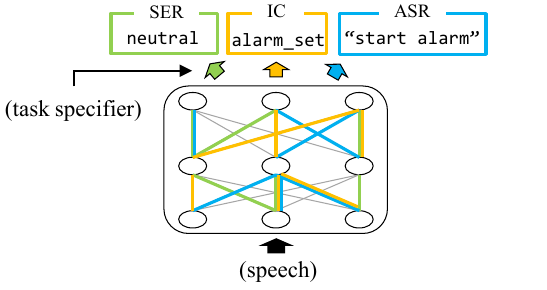}
\caption{Illustration of task-specific subnetworks in multi-task SLU model. To solve SER task, only subnetwork represented as green pathways is activated.}
\label{fig:overview}
\vspace{-5pt}
\end{figure}
During training and inference of a task, only a subnetwork for the task is activated, as shown in Figure \ref{fig:overview}.
While pruning reduces parameter counts, we also expect better continual learning via network pruning, not investigated in \cite{Yang23-LAP}.
During training, only parameters in the task-specific subnetwork are updated while others remain unchanged, which works to mitigate catastrophic forgetting of other tasks.

In this study, we focus on finding task-specific subnetworks in UniverSLU \cite{Arora23-USLU}.
We have trained the model for ER on IEMOCAP corpus \cite{Busso08-IEMOCAP}, IC on SLURP \cite{Bastianelli20-SLURP}, and ASR on LibriSpeech \cite{Panayotov15-LS}.
We found that pruned models achieved better performances on ER and IC with much smaller parameter counts.
We also explore continual learning on additional ASR or IC data.
We observed that, for pruned models, the performances of tasks not currently trained were less deteriorated.
We further observed that the method was effective for UniverSLU trained on more tasks including named entity recognition (NER), and speech command recognition (SCR), and IC on other datasets, which also offers insight into how similarity between tasks lead to overlap between task-specific subnetworks.

\section{Related work}
\subsection{Network pruning}
\label{sec:related-network-pruning}
Network pruning is a technique for removing unnecessary parameters from neural networks, which can reduce model size and computation costs.
Structured pruning remove parameters in groups, while unstructured pruning remove them individually \cite{Davis20-WSNNP}.
This study focuses on unstructured pruning, where it is less effective to achieve computational efficiency in modern libraries and hardware.
However, both often share the same insights, and these insights can be applied to each other, as indicated in previous studies~\cite{Lai21-PARP, Frankle18-LTH}.
The Lottery Ticket Hypothesis (LTH) study \cite{Frankle18-LTH} has demonstrated the existence of sparse subnetworks (called ``winning tickets'') that match or even surpass the performance of the original dense networks.
There are several recent studies on pruning in speech processing \cite{Lai21-PARP, Ding22-AL, Peng23-SP, Yang22-OSD, Yang23-LAP}.
In \cite{Ding22-AL}, LTH for ASR has been investigated.
Pruning of self-supervised learning (SSL) based models have been investigated in \cite{Lai21-PARP} and \cite{Peng23-SP}.
Some other studies focus on obtaining multiple subnetworks of different properties within a single model \cite{Yang22-OSD, Yang23-LAP}.
Omni-sparsity DNN \cite{Yang22-OSD} trains subnetworks of different sparsity.
In \cite{Yang23-LAP}, language-specific subnetworks in a multi-lingual ASR model are considered.
Unlike these studies, our study focuses on a multi-task SLU model and looks for subnetworks for different SLU tasks.
Moreover, our study represents the first investigation into the benefits of continual learning via pruning in SLU.

\subsection{Continual learning}
Continual learning, or lifelong learning, aims to learn new data while preventing catastrophic forgetting of previously learned knowledge.
Continual learning methods can be categorized into three: regularization-based, replay-based, and architecture-based methods.
Regularization-based methods introduce an additional regularization term \cite{Li16-LWF, Kirkpatrick16-EWC}, which has also been investigated in ASR \cite{Chang21-TLL} and SLU \cite{Cappellazzo23-SKDSLU, MYang24-EICLSLU}.
Replay-based methods replay examples from previous data \cite{Lopez-Paz17-GEM}, including studies for ASR \cite{Yang22-OCL}.
Architecture-based methods spare isolated parameters for new task.
This can be done by updating only a part of the entire parameters \cite{Takashima22-UOE, Diwan22-DC, Eeckt23-UA}.
Especially in \cite{Mallya18-PN} and \cite{Golkar19-CLNP}, network pruning is applied sequentially for image recognition tasks, where parameters not used in preceding tasks are allocated for the current task.
In contrast, parameter sharing between tasks can exist in our study.
This is more parameter efficient and flexible, where each subnetwork can be designed with any sparsity.

\section{Finding task-specific subnetworks}
\label{sec:method}
In this study, we propose finding task-specific subnetworks in a multi-task SLU model.
We aim to identify individual pathways for each SLU task on a single dense network of shared parameters, which we call \textit{multi-task pruning}.
On the other hand, obtaining individual sparse models for SLU tasks is called \textit{single-task pruning}.
Multi-task pruning allows us to switch tasks by switching only pathways, which is parameter efficient and preferable for deployment.
To realize this, we adapt a pruning method originally proposed for multi-lingual ASR \cite{Yang23-LAP} to multi-task SLU.

Let $f(X, s_t; \, \theta)$ denote a multi-task SLU network with input speech $X$, prompt of task specifier $s_t$, and parameters $\theta$.
Task specifier is defined for each task $t \in \mathcal{T}$ and instruct the network which task to solve, where $\mathcal{T}$ denotes a set of SLU tasks.
For pruning, we reformulate $f(X, s_t; \theta)$ as $f(X, s_t; m \odot \theta)$, where $m \in \{0, 1\}^{|\theta|}$ denotes a pruning mask.
We assume task-specific pruning, where $m$ is task-specific denoted as $m_t$ for task $t$.
Our multi-task pruning consists of two steps: (1) identifying pruning mask $m_t$ and (2) updating parameters $\theta$ using $m_t$.

In the first step we identify the task-specific pruning mask, as summarized in Algorithm \ref{algo:identify-mask}.
First, we randomly select a task $t$ at the beginning of the loop (Line 3).
Then, we train the network on batches from dataset of task $t$ denoted as $\mathcal{D}_t$ for $N_1$ iterations (Line 5).
After training, pruning is done by setting the pruned position of $m_t$ to $0$ (Line 7).
In this study, we apply global pruning, where the parameters with the top $p\%$ smallest magnitude across all the layers are pruned.
After pruning, the parameters are reset to $\theta_0$ based on LTH \cite{Frankle18-LTH} (Line 8).
We assume iterative pruning, where pruning is repeatedly applied $Q$ times, leading to subnetworks of $1-(1-p)^Q\%$ sparsity.

\begin{figure}[t]
\vspace{-10pt}
\begin{algorithm}[H]
\caption{Identify pruning mask}
\label{algo:identify-mask}
\begin{algorithmic}[1]
\STATE Initialize $f(X, s_t;m_t \odot \theta)$ with $\theta \leftarrow \theta_0$, $m_t \leftarrow \{1\}^{|\theta|}$.
\REPEAT
\FOR{$t \in \mathcal{T}$}
    \REPEAT
    \STATE Update $\theta \leftarrow {\rm TrainNetwork}(f(X,s_t;m_t \odot \theta), \mathcal{D}_t)$
    \UNTIL{The loop repeated $N_1$ times}
    \STATE Update $m_t \leftarrow {\rm Prune}(\theta, m_t, p)$
    \STATE Reset $\theta \leftarrow \theta_0$.
\ENDFOR
\UNTIL{The loop repeated $Q$ times}
\end{algorithmic}
\end{algorithm}
\vspace{-20pt}
\end{figure}

\begin{figure}[t]
\vspace{-10pt}
\begin{algorithm}[H]
\caption{Update parameters}
\label{algo:update-param}
\begin{algorithmic}[1]
\STATE Initialize $f(X,s_t;m_t \odot \theta)$ with $\theta \leftarrow \theta_0$, $m_t$ from Algo.\ref{algo:identify-mask}.
\REPEAT
\FOR{$t \in \mathcal{T}$}
    \REPEAT
        \STATE Update $\theta \leftarrow {\rm TrainNetwork}(f(X,s_t;m_t \odot \theta), \mathcal{D}_t)$
    \UNTIL{The loop repeated $N_2$ times}
\ENDFOR
\UNTIL{The loop repeated $R$ times}
\end{algorithmic}
\end{algorithm}
\vspace{-20pt}
\end{figure}

Consequently, we update the parameters based on the identified mask $m_t$, as in Algorithm \ref{algo:update-param}.
Model parameters $\theta$ are shared across tasks, which corresponds to multi-task training.
Note that each task has its own subnetwork using $m_t$ denoted as $m_t \odot \theta$, so the parameters at subnetwork level are different between tasks.
We randomly select task $t$ (Line 3) and train the network on $\mathcal{D}_t$ (Line 5).
The training is done for $N_2$ iterations, where we set a small value ($N_1 \gg N_2$) for the model not to be biased toward the lately trained tasks.
In single-task pruning for comparison, Algorithm \ref{algo:update-param} is performed independently for each task by employing $\theta_t$.

\begin{figure*}[t]
\centering
\subfloat[LibriSpeech ASR]{\includegraphics[width=1.9\columnwidth]{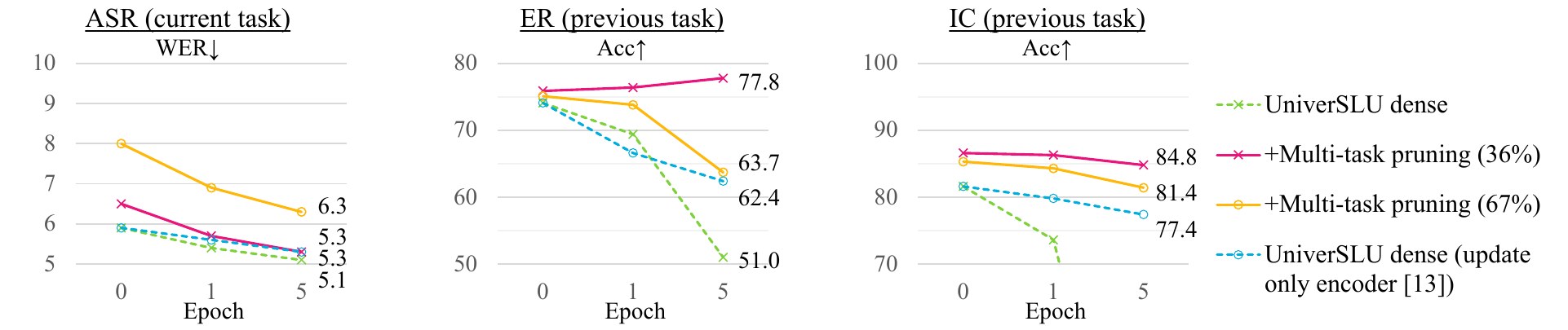}
\label{fig:CL-LIBRI}}
\\
\subfloat[SLURP IC]{\includegraphics[width=1.9\columnwidth]{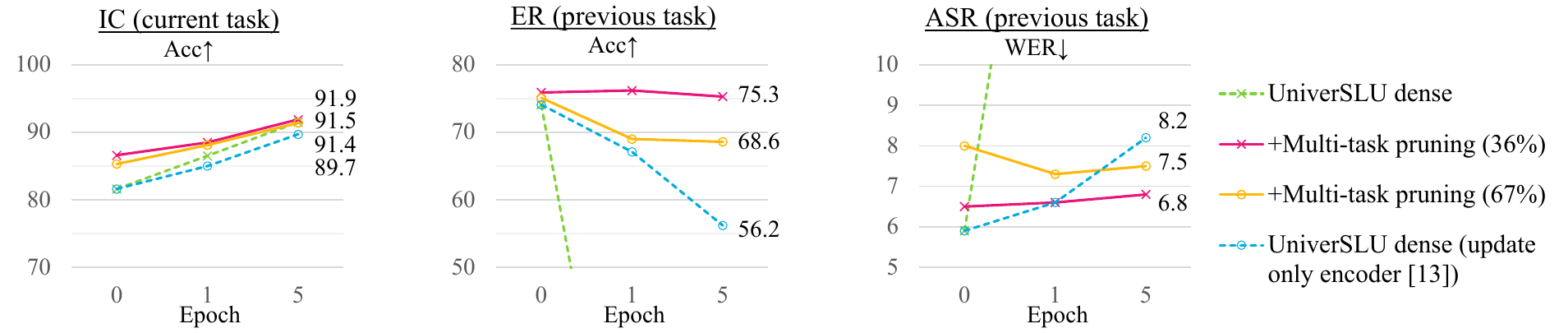}
\label{fig:CL-SLURP}}
\vspace{-5pt}
\caption{Continual learning on (a) LibriSpeech ASR and (b) SLURP IC. We additionally trained models on LibriSpeech $360$h or SLURP real+synthetic data.}
\label{fig:CL}
\vspace{-10pt}
\end{figure*}

Our multi-task pruning not only brings model compression but also has the ability of continual learning, which has not been investigated in \cite{Yang23-LAP}.
We assume the case that, when the new data of task $t_{\rm c}$ become available, we additionally train the model only on the task $t_{\rm c}$ data.
As continual learning in this study, we aim to improve the performance of task $t_{\rm c}$, without degrading the performances of other tasks $t \neq t_{\rm c}$.
For the pruned models, we update only a portion of parameters of a task-specific subnetwork $\{ \theta_j | m_{t_c,j} = 1 \}$ for continual learning.
Other parameters $\{ \theta_j | m_{t_c,j} = 0 \}$ remain unchanged, some of which is used in a subnetwork for previously trained tasks $t \neq t_{\rm c}$.
This can play an important role in retaining the knowledge of such tasks.
Finally, our pruning provides some interpretability, by examining the identified structures of task-specifics subnetworks and their overlap across tasks.

\begingroup
\renewcommand{\arraystretch}{1.1}
\begin{table}[t]
\caption{Performances of dense and pruned models. Param(\%) indicates percentage of nonzero parameters to perform single task (One) or all tasks (All). Multi-task pruning (our proposal) leads to different subnetworks in shared model, while single-task pruning leads to different models. Task-agnostic pruning means single subnetwork for all task.}
\label{tab:main}
\centering
\begin{tabular}{lcccc} \hline
 & Param(\%) & ER & IC & ASR \\
 & One / All & Acc$\uparrow$ & Acc$\uparrow$ & WER$\downarrow$ \\ \hline
UniverSLU dense & $100$ /$100$ & $74.1$ & $81.6$ & $5.9$ \\ \hline
\multicolumn{2}{l}{{\it +Task-specific pruning (${\it 36\%}$)}} \\
\underline{Multi-task} & $64.1$/ $71.0$ & $75.9$ & $86.6$ & $6.5$ \\
Single-task & $64.1$/$192.3$ & $77.0$ & $86.7$ & $6.4$ \\ \hline
+Task-agnostic & $64.1$/ $64.1$ & $74.2$ & $84.7$ & $6.9$ \\ \hline
\multicolumn{2}{l}{{\it +Task-specific pruning (${\it 67\%}$)}} \\
\underline{Multi-task} & $32.9$/ $39.8$ & $75.1$ & $85.3$ & $8.0$ \\
Single-task & $32.9$/ $98.6$ & $75.5$ & $85.2$ & $7.9$ \\ \hline
+Task-agnostic & $32.9$/ $32.9$ & $74.3$ & $84.3$ & $8.3$ \\ \hline
\end{tabular}
\vspace{-10pt}
\end{table}
\endgroup

\begingroup
\renewcommand{\arraystretch}{1.1}
\begin{table*}[t]
\caption{Performances of dense and pruned models for $7$ tasks.}
\label{tab:main7}
\centering
\begin{tabular}{lcccccccc} \hline
 & Param(\%) & ER & IC-SLURP & ASR & IC-FSC & IC-SNIPS & NER & SCR \\
 & One / All & Acc$\uparrow$ & Acc$\uparrow$ & WER$\downarrow$ & Acc$\uparrow$ & F1$\uparrow$ & SLU F1$\uparrow$ & Acc$\uparrow$ \\ \hline
UniverSLU dense & $100$ / $100$ & $75.7$ & $85.1$ & $5.9$ & $99.8$ & $93.1$ & $69.4$ & $98.6$ \\ \hline
+Multi-task pruning & $32.9$ /  $43.7$ & $76.4$ & $84.7$ & $8.1$ & $99.8$ & $94.1$ & $68.4$ & $98.8$ \\
+Single-task pruning & $32.9$ / $230.3$ & $74.7$ & $83.9$ & $8.0$ & $99.7$ & $92.0$ & $68.0$ & $98.7$ \\ \hline
\end{tabular}
\vspace{-15pt}
\end{table*}
\endgroup

\section{Experimental evaluations}

We conducted experiments using UniverSLU \cite{Arora23-USLU} as the multi-task SLU model, obtained by fine-tuning Whisper \cite{Radford22-Whisper} on SLU tasks.
We used Whisper medium of around $760$M parameters as the pre-trained model.
Our initial investigation is based on three datasets: IEMOCAP \cite{Busso08-IEMOCAP}, SLURP \cite{Bastianelli20-SLURP}, and LibriSpeech \cite{Panayotov15-LS}, which will be extended to $7$ tasks in Section \ref{sec:more-tasks}.
IEMOCAP was used for ER, which consists of $12$ hours of speech, and $4$ emotion classes (neutral, happy, sad, and angry) were used in this study.
SLURP was used for IC, which consists of $58$ hours of real and $43.5$ hours of synthetic speech with $69$ intent classes.
We left the synthetic subset for continual learning.
For ASR, we used LibriSpeech the $100$h subset in training and test-other set in evaluation.
The training of UniverSLU was done on the mixture of the above three datasets, where IEMOCAP was upsampled by $10$ times to account for data imbalance.
Following the UniverSLU paper \cite{Arora23-USLU}, we specified which language, task, and dataset to solve by using Whisper-style prompts such as ``\url{<EN>} \url{<IC>} \url{<SLURP>}''.
We conducted the experiments using the ESPnet toolkit \cite{Watanabe18-ESPnet, Arora22-ESPnetSLU} and followed its data pre-processing.

First, we trained a dense UniverSLU model for the three SLU tasks.
Then, we applied network pruning to find task-specific subnetworks in the dense model as in Algorithm \ref{algo:identify-mask} and \ref{algo:update-param}.
Algorithm \ref{algo:identify-mask} iteratively prunes the network by $p=20\%$ for $Q = 2$ or $5$ times, resulting in approximately $36\%$ and $67\%$ sparsity.
We set pruning interval step $N_1$ equivalent to $3$ epochs (average $N_1 = 1300$), where the learning rate was set to $2.0 \times 10^{-4}$ with $2500$ warmup steps.
Global pruning was applied to all the layers of both the Whisper encoder and decoder, except for positional embeddings.
Algorithm \ref{algo:update-param} updates model parameters, switching tasks by average $N_2 = 50$ iterations and repeating $R = 200$ times.
The learning rate was set to $1.0 \times 10^{-5}$ with $1500$ warmup steps.

Table \ref{tab:main} shows the performances of dense and pruned models with different methods.
``Param(\%)'' denotes what percentage of parameters of the dense model remains, or are not zero.
We note what percentages are needed to perform a single task (denoted as ``One'') and all three tasks (denoted as ``All'').
We compared multi-task and single-task pruning described in Section \ref{sec:method}.
Multi-task pruning leads to different pathways $m_t$ within a single model $\theta$, while single-task pruning leads to different models ($\theta_t$ with $m_t$).
To perform a single task, the parameter usage is the same.
To perform all the tasks, for multi-task pruning, the parameter usage is guaranteed not to exceed $100\%$ and was $71.0\%$ for the $36$\% pruned model, due to shared parameters across tasks.
However, for single-task pruning, the parameter usage was $64.1 \times 3 \,{\rm (tasks)} = 193\%$, which is not parameter efficient.
The issue becomes more significant as the number of tasks increases, as seen in Section \ref{sec:more-tasks}.
We also compared these two task-specific pruning with task-agnostic pruning, where pruning is done without distinction of tasks, leading to a single pruned model for all the tasks.
In terms of the task performances in Table \ref{tab:main}, we found that the accuracy of ER and IC was even improved via pruning.
However, we observed performance degradation on ASR, especially for $67$\% pruning.
This would be because ASR is a sequence generation task that requires more parameters than classification tasks ER and IC.
We also observed that multi-task pruning achieved competitive performances against single-task pruning, in a parameter efficient way.
In case of task-agnostic pruning, all the performances lagged behind those of task-specific pruning.
We observed the difference was statistically significant ($p<0.001$) for IC and ASR, using the Matched Pair test.

\subsection{Continual learning}
We also investigate how our models work in continual learning settings, assuming the case that new training data for a specific task becomes available.
We compared training from the dense UniverSLU model and the multi-task pruned models.
We also compared them with updating only the linear layers of encoders in the dense model, which is known as a simple yet effective continual learning method for ASR \cite{Takashima22-UOE}.
We conducted two experiments, using LibriSpeech $360$h (ASR) or SLURP synthetic set (IC) as new data.
In SLURP experiments, as it is difficult to train the model only on the synthetic speech, we trained them on the mixture of real and synthetic SLURP.

Figure \ref{fig:CL-LIBRI} shows the results of the LibriSpeech experiments.
As training went on, the ASR performance got improved for all the models, due to additional ASR training data.
On the other hand, the ER and IC performances largely deteriorated for the dense model (noted as green lines), known as catastrophic forgetting.
The model sometimes outputted ASR results (transcripts), even when it is prompted to perform ER or IC.
The problem was mitigated by updating only its encoder (noted as blue lines).
For pruned models (noted as yellow and red lines), the performance degradation on ER and IC was smaller compared to the dense model and prior continual learning method.
Since only the parameters of the ASR-specific subnetwork are updated during the training, we hypothesize that the remaining parameters can retain the knowledge of other tasks, thus mitigating the issue of catastrophic forgetting.
Remarkably, the $36\%$ pruned model even demonstrated an improvement on ER, despite ER not being trained in the continual learning stage.
This showcases that shared parameters have the potential to extract features beneficial across tasks.

Figure \ref{fig:CL-SLURP} shows the results of the SLURP experiments.
The IC performance was improved by additional data, and the $36\%$ pruned model performed better than the dense model.
Similar to LibriSpeech, the performances of ER and ASR were kept better for pruned models, compared to the dense model and prior continual learning method of updating only encoders.

\subsection{Extending tasks from 3 to 7}
\label{sec:more-tasks}
We add IC on Fluent SC (FSC) \cite{Lugosch19-FSC}, IC on SNIPS \cite{Saade18-SNIPS}, named entity recognition (NER) on SLURP \cite{Bastianelli20-SLURP}, and speech command recognition (SCR) on Google Speech Commands \cite{Warden18-SC} to the training of UniverSLU.
SNIPS and FSC consist of $1.6$K and $30$K utterances with $6$ and $24$ intent classes, respectively.
SLURP also has annotations for NER of $55$ classes.
NER is done by predicting entity tags and corresponding lexical fillers alternately, like ``\url{<entity:date>} \url{<FILL>} \url{tomorrow} \url{<SEP>} ...'', similar to \cite{Arora22-ESPnetSLU}.
It is evaluated on the SLU-F1 metric introduced in \cite{Bastianelli20-SLURP}.
Google Speech Commands (v0.02) contains $36$K utterances for $12$ different commands.
We regard IC on different datasets as different tasks, as intent labels are different, which results in $7$ tasks.

Table \ref{tab:main7} shows the performance of the $7$-task UniverSLU.
Similar to Table \ref{tab:main}, pruning was able to reduce the parameter counts with small performance losses or improvements on some tasks, except for ASR.
Also, multi-task pruning was parameter efficient to solve multiple tasks and achieved consistently better performances than single-task pruning except for the ASR task.

\begin{figure}[t]
\centering
\includegraphics[width=0.85\columnwidth]{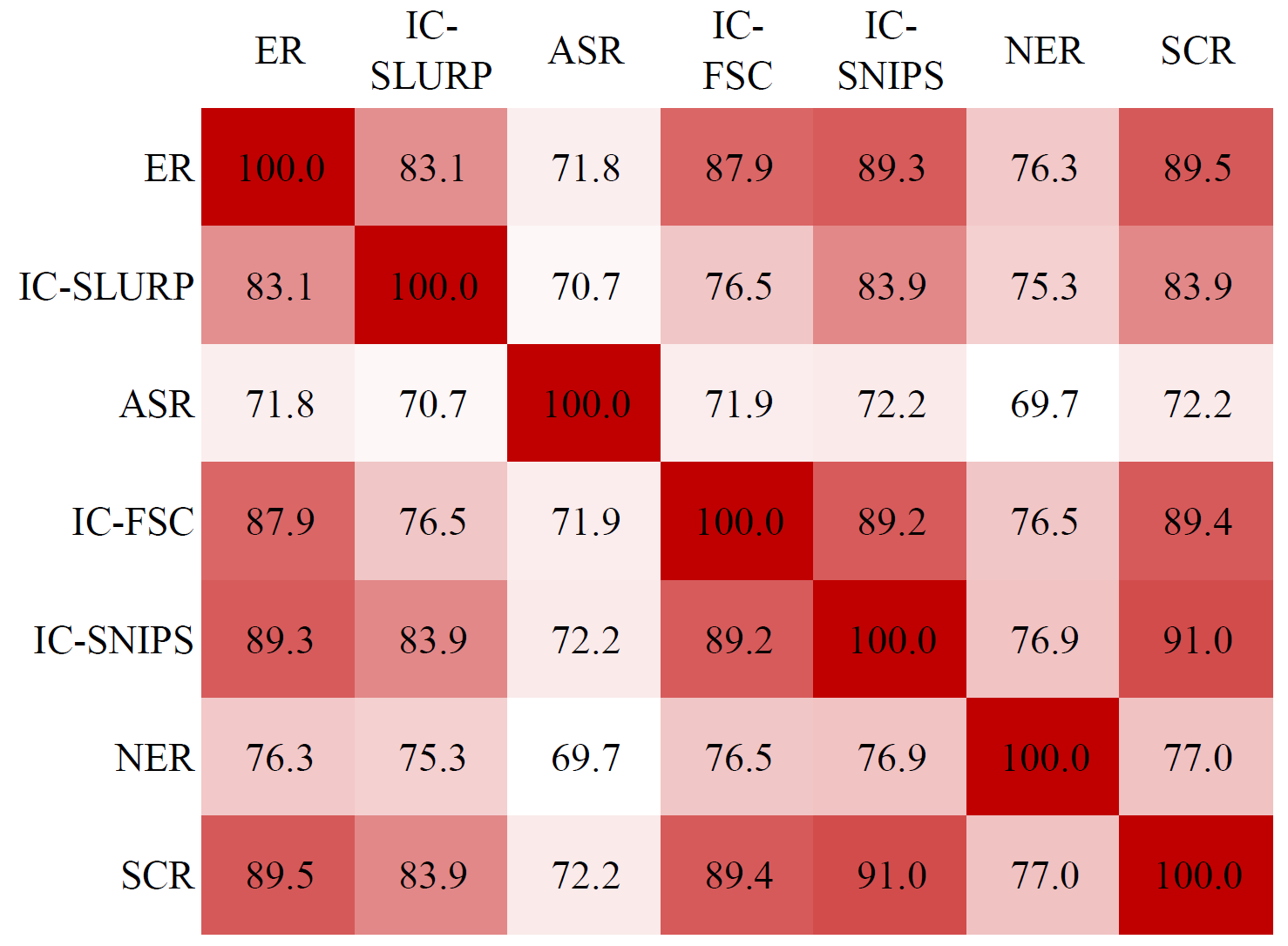}
\caption{Parameter overlap ratio between tasks.}
\label{fig:overlap}
\vspace{-15pt}
\end{figure}

Finally, Figure \ref{fig:overlap} analyzes the difference in pruning masks between tasks.
We calculated the parameter overlap ratio between subnetworks for task $i$ and $j$, following \cite{Yang23-LAP}, as:
${\rm Overlap}(m_i, m_j) = \frac{| m_i = 1 \cap m_j = 1 |}{| m_i = 1 \cup m_j = 1 |}.$
ASR and NER are sequence generation tasks, while others are classification tasks.
This can be the reason why ASR and NER have less overlap with others.
Among classification tasks, the overlap ratios are relatively high, where the overlap between IC-SNIPS and SCR is the highest.
NER performs generation of entity tags along with lexical fillers, which can make its subnetwork also different from ASR.

\vspace{-5pt}
\section{Conclusions}
We have investigated network pruning to obtain task-specific subnetworks within a multi-task SLU model.
We conducted experimental evaluations based on UniverSLU model that covers ER, IC, and ASR.
We found that subnetworks achieved better performances on ER and IC than the dense network, even with $67\%$ sparsity.
In addition to model compression, our approach also has continual learning capabilities.
We also found that, with additional ASR training, the ASR performance can be improved without largely degrading the previously trained ER and IC performances.
As future work, we plan to extend this study by incorporating structured pruning, as discussed in Section \ref{sec:related-network-pruning}.

\bibliographystyle{IEEEtran}
\bibliography{mybib}

\end{document}